\documentclass[11pt]{article}
\usepackage{coling08}
\usepackage{times}
\usepackage{latexsym}
\usepackage{multirow}
\newcommand{\captionfonts}{\small}
\makeatletter  
\long\def\@makecaption#1#2{%
  \vskip\abovecaptionskip
  \sbox\@tempboxa{{\captionfonts #1: #2}}%
  \ifdim \wd\@tempboxa >\hsize
    {\captionfonts #1: #2\par}
  \else
    \hbox to\hsize{\hfil\box\@tempboxa\hfil}%
  \fi
  \vskip\belowcaptionskip}
\makeatother   

\title{A Data Driven Approach to Query Expansion in Question Answering}

\author{Leon Derczynski, Jun Wang, Robert Gaizauskas and Mark A. Greenwood\\
  Department of Computer Science\\
  University of Sheffield\\
  Regent Court, 211 Portobello\\
  Sheffield S1 4DP UK\\
  {\tt \{aca00lad, acp07jw, r.gaizauskas, m.greenwood\}@shef.ac.uk}}

\date{}

\begin{document}
\maketitle

\begin{abstract}
Automated answering of natural language questions is an interesting and useful problem to solve. Question answering (QA) systems often perform information retrieval at an initial stage. Information retrieval (IR) performance, provided by engines such as Lucene, places a bound on overall system performance. For example, no answer bearing documents are retrieved at low ranks for almost 40\% of questions. 

In this paper, answer texts from previous QA evaluations held as part of the Text REtrieval Conferences (TREC) are paired with queries and analysed in an attempt to identify performance-enhancing words. These words are then used to evaluate the performance of a query expansion method.

Data driven extension words were found to help in over 70\% of difficult questions. These words can be used to improve and evaluate query expansion methods. Simple blind relevance feedback (RF) was correctly predicted as unlikely to help overall performance, and an possible explanation is provided for its low value in IR for QA.

\end{abstract}

\section{Introduction}\label{intro}

The task of supplying an answer to a question, given some background knowledge, is often considered fairly trivial from a human point of view, as long as the question is clear and the answer is known. The aim of an automated question answering system is to provide a single, unambiguous response to a natural language question, given a text collection as a knowledge source, within a certain amount of time. Since 1999, the Text Retrieval Conferences have included a task to evaluate such systems, based on a large pre-defined corpus (such as AQUAINT, containing around a million news articles in English) and a set of unseen questions.

Many information retrieval systems perform document retrieval, giving a list of potentially relevant documents when queried -- Google's and Yahoo!'s search products are examples of this type of application. Users formulate a query using a few keywords that represent the task they are trying to perform; for example, one might search for ``eiffel tower height" to determine how tall the Eiffel tower is. IR engines then return a set of references to potentially relevant documents.

In contrast, QA systems must return an exact answer to the question. They should be confident that the answer has been correctly selected; it is no longer down to the user to research a set of document references in order to discover the information themselves. Further, the system takes a natural language question as input, instead of a few user-selected key terms.

Once a QA system has been provided with a question, its processing steps can be described in three parts - Question Pre-Processing, Text Retrieval and Answer Extraction:

\paragraph{1. Question Pre-Processing} TREC questions are grouped into series which relate to a given target. For example, the target may be ``Hindenburg disaster" with questions such as ``What type of craft was the Hindenburg?" or ``How fast could it travel?". Questions may include pronouns referencing the target or even previous answers, and as such require processing before they are suitable for use. 
\paragraph{2. Text Retrieval} An IR component will return a ranked set of texts, based on query terms. Attempting to understand and extract data from an entire corpus is too resource intensive, and so an IR engine defines a limited subset of the corpus that is likely to contain answers. The question should have been pre-processed correctly for a useful set of texts to be retrieved -- including anaphora resolution.
\paragraph{3. Answer Extraction (AE)} Given knowledge about the question and a set of texts, the AE system attempts to identify answers. It should be clear that only answers within texts returned by the IR component have any chance of being found.
\medskip

Reduced performance at any stage will have a knock-on effect, capping the performance of later stages. If questions are left unprocessed and full of pronouns (e.g.,``When did it sink?") the IR component has very little chance of working correctly -- in this case, the desired action is to retrieve documents related to the Kursk submarine, which would be impossible.

IR performance with a search engine such as Lucene returns no useful documents for at least 35\% of all questions -- when looking at the top 20 returned texts. This caps the AE component at 65\% question ``coverage". We will measure the performance of different IR component configurations, to rule out problems with a default Lucene setup.

For each question, answers are provided in the form of regular expressions that match answer text, and a list of documents containing these answers in a correct context. As references to correct documents are available, it is possible to explore a data-driven approach to query analysis. We determine which questions are hardest then concentrate on identifying helpful terms found in correct documents, with a view to building a system than can automatically extract these helpful terms from unseen questions and supporting corpus. The availability and usefulness of these terms will provide an estimate of performance for query expansion techniques.

There are at least two approaches which could make use of these term sets to perform query expansion. They may occur in terms selected for blind RF (non-blind RF is not applicable to the TREC QA task). It is also possible to build a catalogue of terms known to be useful according to certain question types, thus leading to a dictionary of (known useful) expansions that can be applied to previously unseen questions. We will evaluate and also test blind relevance feedback in IR for QA.

\section{Background and Related Work}
The performance of an IR system can be quantified in many ways. We choose and define measures pertinent to IR for QA. Work has been done on relevance feedback specific to IR for QA, where it is has usually be found to be unhelpful. We outline the methods used in the past, extend them, and provide and test means of validating QA relevance feedback.

\subsection{Measuring QA Performance}
This paper uses two principle measures to describe the performance of the IR component. \emph{Coverage} is defined as the proportion of questions where at least one answer bearing text appears in the retrieved set. \emph{Redundancy} is the average number of answer bearing texts retrieved for each question~\cite{Roberts:Passage}. 

Both these measures have a fixed limit $n$ on the number of texts retrieved by a search engine for a query. As redundancy counts the number of texts containing correct answers, and not instances of the answer itself, it can never be greater than the number of texts retrieved.

The TREC reference answers provide two ways of finding a correct text, with both a regular expression and a document ID. Lenient hits (retrievals of answer bearing documents) are those where the retrieved text matches the regular expression; strict hits occur when the document ID of the retrieved text matches that declared by TREC as correct \emph{and} the text matches the regular expression. Some documents will match the regular expression but not be deemed as containing a correct answer (this is common with numbers and dates~\cite{MIR:166}), in which case a lenient match is found, but not a strict one.

The answer lists as defined by TREC do not include every answer-bearing document -- only those returned by previous systems and marked as correct. Thus, false negatives are a risk, and strict measures place an approximate lower bound on the system's actual performance. Similarly, lenient matches can occur out of context, without a supporting document; performance based on lenient matches can be viewed as an approximate upper bound~\cite{lin2005brt}.

\subsection{Relevance Feedback}
Relevance feedback is a widely explored technique for query expansion. It is often done using a specific measure to select terms using a limited set of ranked documents of size $r$; using a larger set will bring term distribution closer to values over the whole corpus, and away from ones in documents relevant to query terms. Techniques are used to identify phrases relevant to a query topic, in order to reduce noise (such as terms with a low corpus frequency that relate to only a single article) and query drift~\cite{roussinov,allan96incremental}.

In the context of QA, Pizzato~\shortcite{pizzato} employs blind RF using the AQUAINT corpus in an attempt to improve performance when answering factoid questions on personal names. This is a similar approach to some content in this paper, though limited to the study of named entities, and does not attempt to examine extensions from the existing answer data.

Monz~\shortcite{monz2003drq} finds a negative result when applying blind feedback for QA in TREC 9, 10 and 11, and a neutral result for TREC 7 and 8's ad hoc retrieval tasks. Monz's experiment, using $r=10$ and standard Rocchio term weighting, also found a further reduction in performance when $r$ was reduced (from 10 to 5). This is an isolated experiment using just one measure on a limited set of questions, with no use of the available answer texts.

Robertson~\shortcite{robertson92okapi} notes that there are issues when using a whole document for feedback, as opposed to just a single relevant passage; as mentioned in Section~\ref{method:irengines}, passage- and document-level retrieval sets must also be compared for their performance at providing feedback. Critically, we will survey the intersection between words known to be helpful and blind RF terms based on initial retrieval, thus showing exactly how likely an RF method is to succeed.

\section{Methodology}
We first investigated the possibility of an IR-component specific failure leading to impaired coverage by testing a variety of IR engines and configurations. Then, difficult questions were identified, using various performance thresholds. Next, answer bearing texts for these harder questions were checked for words that yielded a performance increase when used for query expansion.
After this, we evaluated how likely a RF-based approach was to succeed. Finally, blind RF was applied to the whole question set. IR performance was measured, and terms used for RF compared to those which had proven to be helpful as extension words.

\subsection{IR Engines}\label{method:irengines}
A QA framework \cite{Greenwood:Answerfinder} was originally used to construct a QA system based on running a default Lucene installation. As this only covers one IR engine in one configuration, it is prudent to examine alternatives. Other IR engines should be tested, using different configurations. The chosen additional engines were: Indri, based on the mature INQUERY engine and the Lemur toolkit~\cite{Lemur}; and Terrier, a newer engine designed to deal with corpora in the terabyte range and to back applications entered into TREC conferences~\cite{terrier}.

We also looked at both passage-level and document-level retrieval. Passages can be defined in a number of ways, such as a sentence, a sliding window of $k$ terms centred on the target term(s), parts of a document of fixed (and equal) lengths, or a paragraph. In this case, the documents in the AQUAINT corpus contain paragraph markers which were used as passage-level boundaries, thus making ``passage-level" and ``paragraph-level" equivalent in this paper. Passage-level retrieval may be preferable for AE, as the number of potential distracters is somewhat reduced when compared to document-level retrieval~\cite{Roberts:Passage}.

The initial IR component configuration was with Lucene indexing the AQUAINT corpus at passage-level, with a Porter stemmer~\cite{Porter:stemming} and an augmented version of the CACM~\cite{jones1976irt} stopword list.


Indri natively supports document-level indexing of TREC format corpora. Passage-level retrieval was done using the paragraph tags defined in the corpus as delimiters; this allows both passage- and document-level retrieval from the same index, according to the query. 

All the IR engines were unified to use the Porter stemmer and the same CACM-derived stopword list.

The top $n$ documents for each question in the TREC2004, TREC2005 and TREC2006 sets were retrieved using every combination of engine, and configuration\footnote{Save Terrier / TREC2004 / passage-level retrieval; passage-level retrieval with Terrier was very slow using our configuration, and could not be reliably performed using the same Terrier instance as document-level retrieval.}. The questions and targets were processed to produce IR queries as per the default configuration for the QA framework. Examining the top 200 documents gave a good compromise between the time taken to run experiments (between 30 and 240 minutes each) and the amount one can mine into the data. Tabulated results are shown in Table~\ref{basepassagelevelperformance} and Table~\ref{basedocumentlevelperformance}. Queries have had anaphora resolution performed in the context of their series by the QA framework. AE components begin to fail due to excess noise when presented with over 20 texts, so this value is enough to encompass typical operating parameters and leave space for discovery~\cite{Greenwood:TREC2006}.

A failure analysis (FA) tool, an early version of which is described by~\cite{Sanka:Passage}, provided reporting and analysis of IR component performance. In this experiment, it provided high level comparison of all engines, measuring coverage and redundancy as the number of documents retrieved, $n$, varies. This is measured because a perfect engine will return the most useful documents first, followed by others; thus, coverage will be higher for that engine with low values of $n$.

\begin{table}
\small
\begin{center}
\begin{tabular}{ | r | r || c | c | c | c | }
\hline
 \multicolumn{2}{ | c | }{} &
 \multicolumn{2}{ | c | }{Coverage} &
 \multicolumn{2}{ | c | }{Redundancy} \\
\hline
 & Year & Len. & Strict & Len. & Strict \\
\hline
\multirow{4}{*}{Lucene} 
 & 2004 & 0.686 & 0.636 & 2.884 & 1.624 \\
 & 2005 & 0.703 & 0.566 & 2.780 & 1.155 \\
 & 2006 & 0.665 & 0.568 & 2.417 & 1.181 \\
\hline
\multirow{4}{*}{Indri} 
 & 2004 & 0.690 & 0.554 & 3.849 & 1.527 \\
 & 2005 & 0.694 & 0.512 & 3.908 & 1.056 \\
 & 2006 & 0.691 & 0.552 & 3.373 & 1.152 \\
\hline
\multirow{4}{*}{Terrier}
 & 2004 & - & - & - & - \\
 & 2005 & - & - & - & - \\
 & 2006 & 0.638 & 0.493 & 2.520 & 1.000 \\
\hline
\end{tabular}
\caption{Performance of Lucene, Indri and Terrier at paragraph level, over top 20 documents. This clearly shows the limitations of the engines.\label{basepassagelevelperformance}}
\end{center}
\end{table}
\normalsize

\begin{table}
\small
\begin{center}
\begin{tabular}{ | r | r || c | c | c | c | }
\hline
 \multicolumn{2}{ | c | }{} &
 \multicolumn{2}{ | c | }{Coverage} &
 \multicolumn{2}{ | c | }{Redundancy} \\
\hline
 & Year & Len. & Strict & Len. & Strict \\
\hline
\multirow{4}{*}{Indri}
 & 2004 & 0.926 & 0.837 & 7.841 & 2.663 \\
 & 2005 & 0.935 & 0.735 & 7.573 & 1.969 \\
 & 2006 & 0.882 & 0.741 & 6.872 & 1.958 \\
\hline
\multirow{4}{*}{Terrier} 
 & 2004 & 0.919 & 0.806 & 7.186 & 2.380 \\
 & 2005 & 0.928 & 0.766 & 7.620 & 2.130 \\
 & 2006 & 0.983 & 0.783 & 6.339 & 2.067 \\
\hline
\end{tabular}
\caption{Performance of Indri and Terrier at document level IR over the AQUAINT corpus, with $n=20$\label{basedocumentlevelperformance}}
\end{center}
\end{table}
\normalsize

\subsection{Identification of Difficult Questions}

Once the performance of an IR configuration over a question set is known, it's possible to produce a simple report listing redundancy for each question. A performance reporting script accesses the FA tool's database and lists all the questions in a particular set with the strict and lenient redundancy for selected engines and configurations. Engines may use passage- or document-level configurations. 


Data on the performance of the three engines is described in Table~\ref{basedocumentlevelperformance}. As can be seen, the coverage with passage-level retrieval (which was often favoured, as the AE component performs best with reduced amounts of text) languishes between 51\% and 71\%, depending on the measurement method. Failed anaphora resolution may contribute to this figure, though no deficiencies were found upon visual inspection.

Not all documents containing answers are noted, only those checked by the NIST judges~\cite{bilottiadvice}. Match judgements are incomplete, leading to the potential generation of false negatives, where a correct answer is found with complete supporting information, but as the information has not been manually flagged, the system will mark this as a failure. Assessment methods are fully detailed in Dang et al.~\shortcite{dang2006otq}. Factoid performance is still relatively poor, although as only 1.95 documents match per question, this may be an effect of such false negatives~\cite{Voorhees:TREC12}. Work has been done into creating synthetic corpora that include exhaustive answer sets~\cite{bilotti2004qet,tellex2003qep,lin2005brt}, but for the sake of consistency, and easy comparison with both parallel work and prior local results, the TREC judgements will be used to evaluate systems in this paper.

Mean redundancy is also calculated for a number of IR engines. Difficult questions were those for which no answer bearing texts were found by either strict or lenient matches in any of the top $n$ documents, using a variety of engines. As soon as one answer bearing document was found by an engine using any measure, that question was deemed \emph{non-difficult}.  Questions with mean redundancy of zero are marked \emph{difficult}, and subjected to further analysis. Reducing the question set to just difficult questions produces a TREC-format file for re-testing the IR component.

\subsection{Extension of Difficult Questions}
The documents deemed relevant by TREC must contain some useful text that can help IR engine performance. Such words should be revealed by a gain in redundancy when used to extend an initially difficult query, usually signified by a change from zero to a non-zero value (signifying that relevant documents have been found where none were before). In an attempt to identify where the useful text is, the relevant documents for each difficult question were retrieved, and passages matching the answer regular expression identified. A script is then used to build a list of terms from each passage, removing words in the question or its target, words that occur in the answer, and stopwords (based on both the indexing stopword list, and a set of stems common within the corpus). In later runs, numbers are also stripped out of the term list, as their value is just as often confusing as useful~\cite{MIR:166}. Of course, answer terms provide an obvious advantage that would not be reproducible for questions where the answer is unknown, and one of our goals is to help query expansion for unseen questions. This approach may provide insights that will enable appropriate query expansion where answers are not known.

Performance has been measured with both the question followed by an extension (Q+E), as well as the question followed by the target and then extension candidates (Q+T+E). Runs were also executed with just Q and Q+T, to provide non-extended reference performance data points. Addition of the target often leads to gains in performance~\cite{arizonasu}, and may also aid in cases where anaphora resolution has failed.

Some words are retained, such as titles, as including these can be inferred from question or target terms and they will not unfairly boost redundancy scores; for example, when searching for a ``Who" question containing the word ``military", one may want to preserve appellations such as ``Lt." or ``Col.", even if this term appears in the answer.

This filtered list of extensions is then used to create a revised query file, containing the base question (with and without the target suffixed) as well as new questions created by appending a candidate extension word.

Results of retrievals with these new question are loaded into the FA database and a report describing any performance changes is generated. The extension generation process also creates custom answer specifications, which replicate the information found in the answers defined by TREC. 

This whole process can be repeated with varying question difficulty thresholds, as well as alternative $n$ values (typically from 5 to 100), different engines, and various question sets.

\subsection{Relevance Feedback Performance}\label{rfmethod}
Now that we can find the helpful extension words (HEWs) described earlier, we're equipped to evaluate query expansion methods. One simplistic approach could use blind RF to determine candidate extensions, and be considered potentially successful should these words be found in the set of HEWs for a query. For this, term frequencies can be measured given the top $r$ documents retrieved using anaphora-resolved query $Q$. After stopword and question word removal, frequent terms are appended to $Q$, which is then re-evaluated. This has been previously attempted for factoid questions~\cite{arizonasu} and with a limited range of $r$ values~\cite{monz2003drq} but not validated using a set of data-driven terms.

We investigated how likely term frequency (TF) based RF is to discover HEWs. To do this, the proportion of HEWs that occurred in initially retrieved texts was measured, as well as the proportion of these texts containing at least one HEW. Also, to see how effective an expansion method is, suggested expansion terms can be checked against the HEW list.

We used both the top 5 and the top 50 documents in formulation of extension terms, with TF as a ranking measure; 50 is significantly larger than the optimal number of documents for AE (20), without overly diluting term frequencies.

Problems have been found with using entire documents for RF, as the topic may not be the same throughout the entire discourse~\cite{robertson92okapi}. Limiting the texts used for RF to paragraphs may reduce noise; both document- and paragraph-level terms should be checked.

\section{Results}

Once we have HEWs, we can determine if these are going to be of significant help when chosen as query extensions. We can also determine if a query expansion method is likely to be fruitful. Blind RF was applied, and assessed using the helpful words list, as well as RF's effect on coverage.

\begin{table}
\small
\begin{center}
\begin{tabular}{ | r || c | c | c | c | }
\hline
 & \multicolumn{4}{ | c | }{Engine} \\
\hline
 Year & \parbox[t][0.24in]{0.4in}{Lucene\\Para} & \parbox[t][0.24in]{0.4in}{Indri\\Para} & \parbox[t][0.24in]{0.4in}{Indri\\Doc} & \parbox[t][0.24in]{0.4in}{Terrier\\Doc} \\
\hline
 2004 & 76 & 72 & 37 & 42 \\
 2005 & 87 & 98 & 37 & 35 \\
 2006 & 108 & 118 & 59 & 53 \\
\hline
\end{tabular}
\caption{Number of difficult questions, as defined by those which have zero redundancy over both strict and lenient measures, at $n=20$. Questions seem to get harder each year. Document retrieval yields fewer difficult questions, as more text is returned for potential matching.\label{difficultquestioncounts}}
\end{center}
\end{table}
\normalsize

\begin{table}
\small
\begin{center}
\begin{tabular}{ | l | c | c | c | }
 \multicolumn{4}{ c }{Engine} \\
\hline
 & Lucene & Indri & Terrier \\
\hline
 Paragraph & 226 & 221 & - \\
 Document & - & 121 & 109 \\
\hline
\end{tabular}
\caption{Number of difficult questions in the 2006 task, as defined above, this time with $n=5$. Questions become harder as fewer chances are given to provide relevant documents.\label{difficultquestioncounts2006n5}}
\end{center}
\end{table}
\normalsize

\subsection{Difficult Question Analysis}

\begin{table}
\small
\begin{center}
\begin{tabular}{ | r | r || c | c | }
\hline
\multicolumn{2}{ | c | }{\multirow{2}{*}{}}
 & \multicolumn{2}{ | c | }{Match type} \\
\hline
 & & Strict & Lenient \\
\hline
\multirow{3}{*}{Year}
 & 2004 & 39 & 49 \\
 & 2005 & 56 & 66 \\
 & 2006 & 53 & 49 \\
\hline
\end{tabular}
\caption{Common difficult questions (over all three engines mentioned above) by year and match type; $n=20$.\label{dqstrictlenient}}
\end{center}
\end{table}
\normalsize

\begin{table}
\small
\begin{center}
\begin{tabular}{ | l | r | }
\hline
Difficult questions used & 118 \\
Variations tested & 6683 \\
Questions that benefited & 87 (74.4\%) \\
Helpful extension words (strict) & 4973 \\
Mean helpful words per question & 42.144 \\
Mean redundancy increase & 3.958 \\
\hline
\end{tabular}
\caption{Using Terrier Passage / strict matching, retrieving 20 docs, with TREC2006 questions / AQUAINT. Difficult questions are those where no strict matches are found in the top 20 IRT from just one engine.\label{dq2k6perf}}
\end{center}
\end{table}
\normalsize

The number of difficult questions found at $n=20$ is shown in Table~\ref{difficultquestioncounts}. Document-level retrieval gave many fewer difficult questions, as the amount of text retrieved gave a higher chance of finding lenient matches. A comparison of strict and lenient matching is in Table~\ref{dqstrictlenient}. 


Extensions were then applied to difficult questions, with or without the target. The performance of these extensions is shown in Table~\ref{dq2k6perf}. Results show a significant proportion (74.4\%) of difficult questions can benefit from being extended with non-answer words found in answer bearing texts.

\begin{table}
\small
\begin{center}
\begin{tabular}{ | l || r | r | r |}
\hline
 & 2004 & 2005 & 2006 \\
\hline
HEW found in IRT & 4.17\% & 18.58\% & 8.94\% \\
IRT containing HEW & 10.00\% & 33.33\% & 34.29\% \\
\hline
RF words in HEW & 1.25\% & 1.67\% & 5.71\% \\
\hline
\end{tabular}
\caption{``Helpful extension words": the set of extensions that, when added to the query, move redundancy above zero. $r=5$, $n=20$, using Indri at passage level.\label{hewrfintersection}}
\end{center}
\end{table}
\normalsize

\begin{table}
\small
\begin{center}
\begin{tabular}{ | c || l | l | l | l || l |}
\hline
 & \multicolumn{4}{ | c | }{$r$} & \\
\cline{2-5}
 & \multicolumn{2}{ | c | }{5} & \multicolumn{2}{ | c | }{50} & Baseline \\
\cline{1-5}
Rank & Doc & Para & Doc & Para & \\
\hline
5 & 0.253 & 0.251 & 0.240 & 0.179 & 0.312 \\
10 & 0.331 & 0.347 & 0.331 & 0.284 & 0.434 \\
20 & 0.438 & 0.444 & 0.438 & 0.398 & 0.553 \\
50 & 0.583 & 0.577 & 0.577 & 0.552 & 0.634 \\
\hline
\end{tabular}
\caption{Coverage (strict) using blind RF. Both document- and paragraph-level retrieval used to determine RF terms.\label{brf2k6}}
\end{center}
\end{table}
\normalsize

\subsection{Applying Relevance Feedback}
Identifying HEWs provides a set of words that are useful for evaluating potential expansion terms. Using simple TF based feedback (see Section~\ref{rfmethod}), 5 terms were chosen per query. These words had some intersection (see Table~\ref{hewrfintersection}) with the extension words set, indicating that this RF may lead to performance increases for previously unseen questions. Only a small number of the HEWs occur in the initially retrieved texts (IRTs), although a noticeable proportion of IRTs (up to 34.29\%) contain at least one HEW. However, these terms are probably not very frequent in the documents and unlikely to be selected with TF-based blind RF. The mean proportion of RF selected terms that were HEWs was only 2.88\%. Blind RF for question answering fails here due to this low proportion. Strict measures are used for evaluation as we are interested in finding documents which were not previously being retrieved rather than changes in the distribution of keywords in IRT.

Document and passage based RF term selection is used, to explore the effect of noise on terms, and document based term selection proved marginally superior. Choosing RF terms from a small set of documents ($r=5$) was found to be marginally better than choosing from a larger set ($r=50$). In support of the suggestion that RF would be unlikely to locate HEWs, applying blind RF consistently hampered overall coverage (Table~\ref{brf2k6}). 

\section{Discussion}
\begin{table}
\small
\begin{center}
\begin{tabular}{ | l | r | }
\hline
  \multicolumn{2}{ | l | }{\emph{Question:}} \\
  \multicolumn{2}{ | l | }{Who was the nominal leader after the overthrow?} \\
\hline
  \multicolumn{2}{ | l | }{\emph{Target:} Pakistani government overthrown in 1999} \\
\hline
  Extension word & Redundancy \\
\hline
  Kashmir & 4 \\
  Pakistan & 4 \\
  Islamabad & 2.5 \\
\hline
\hline
  \multicolumn{2}{ | l | }{\emph{Question:} Where did he play in college?} \\
\hline
  \multicolumn{2}{ | l | }{\emph{Target:} Warren Moon} \\
\hline
  Extension word & Redundancy \\
\hline
  NFL & 2.5 \\
  football & 1 \\
\hline
\hline
  \multicolumn{2}{ | l | }{\emph{Question:} Who have commanded the division?} \\
\hline
  \multicolumn{2}{ | l | }{\emph{Target:} 82nd Airborne division} \\
\hline
  Extension word & Redundancy \\
\hline
 Gen & 3 \\
 Col & 2 \\
 decimated & 2 \\
 officer & 1 \\
\hline
\end{tabular}
\end{center}
\caption{Queries with extensions, and their mean redundancy using Indri at document level with $n=20$. Without extensions, redundancy is zero.\label{pakistanextensions}}
\end{table}
\normalsize

HEWs are often found in answer bearing texts, though these are hard to identify through simple TF-based RF. A majority of difficult questions can be made accessible through addition of HEWs present in answer bearing texts, and work to determine a relationship between words found in initial retrieval and these HEWs can lead to coverage increases. HEWs also provide an effective means of evaluating other RF methods, which can be developed into a generic rapid testing tool for query expansion techniques. TF-based RF, while finding some HEWs, is not effective at discovering extensions, and reduces overall IR performance.

There was not a large performance change between engines and configurations. Strict paragraph-level coverage never topped 65\%, leaving a significant number of questions where no useful information could be provided for AE.

The original sets of difficult questions for individual engines were small -- often less than the 35\% suggested when looking at the coverage figures. Possible causes could include:

{\bf Difficult questions being defined as those for which average redundancy is zero:} This limit may be too low. To remedy this, we could increase the redundancy limit to specify an arbitrary number of difficult questions out of the whole set.\\
{\bf The use of both strict and lenient measures:} It is possible to get a lenient match (thus marking a question as non-difficult) when the answer text occurs out of context.
\medskip

Reducing $n$ from 20 to 5 (Table~\ref{difficultquestioncounts2006n5}) increased the number of difficult questions produced. From this we can hypothesise that although many search engines are succeeding in returning useful documents (where available), the distribution of these documents over the available ranks is not one that bunches high ranking documents up as those immediately retrieved (unlike a perfect engine; see Section~\ref{method:irengines}), but rather suggests a more even distribution of such documents over the returned set.


The number of candidate extension words for queries (even after filtering) is often in the range of hundreds to thousands. Each of these words creates a separate query, and there are two variations, depending on whether the target is included in the search terms or not. Thus, a large number of extended queries need to be executed for each question run. Passage-level retrieval returns less text, which has two advantages: firstly, it reduces the scope for false positives in lenient matching; secondly, it is easier to scan result by eye and determine why the engine selected a result.


Proper nouns are often helpful as extensions. We noticed that these cropped up fairly regularly for some kinds of question (e.g. ``Who"). Especially useful were proper nouns associated with locations - for example, adding ``Pakistani" to a query containing the word Pakistan lifted redundancy above zero for a question on President Musharraf, as in Table~\ref{pakistanextensions}. This reconfirms work done by Greenwood~\shortcite{Greenwood:pertainyms}.

\section{Conclusion and Future Work}

IR engines find some questions very difficult and consistently fail to retrieve useful texts even with high values of $n$. This behaviour is common over many engines. Paragraph level retrieval seems to give a better idea of which questions are hardest, although the possibility of false negatives is present from answer lists and anaphora resolution.

Relationships exist between query words and helpful words from answer documents (e.g. with a military leadership themes in a query, adding the term ``general" or ``gen" helps). Identification of HEWs has potential use in query expansion. They could be used to evaluate RF approaches, or associated with question words and used as extensions. 

Previous work has ruled out relevance feedback in particular circumstances using a single ranking measure, though this has not been based on analysis of answer bearing texts. The presence of HEWs in IRT for difficult questions shows that guided RF may work, but this will be difficult to pursue. Blind RF based on term frequencies does not increase IR performance. However, there is an intersection between words in initially retrieved texts and words data driven analysis defines as helpful, showing promise for alternative RF methods (e.g. based on TFIDF). These extension words form a basis for indicating the usefulness of RF and query expansion techniques.

In this paper, we have chosen to explore only one branch of query expansion. An alternative data driven approach would be to build associations between recurrently useful terms given question content. Question texts could be stripped of stopwords and proper nouns, and a list of HEWs associated with each remaining term. To reduce noise, the number of times a particular extension has helped a word would be counted. Given sufficient sample data, this would provide a reference body of HEWs to be used as an aid to query expansion.

\bibliographystyle{coling}
\bibliography{ddaqeqa}

\end{document}